\documentclass[conference]{IEEEtran}
\IEEEoverridecommandlockouts
\usepackage{cite}
\usepackage{amsmath,amssymb,amsfonts}
\usepackage{algorithmic}
\usepackage{graphicx}
\usepackage{textcomp}
\usepackage{xcolor}
\usepackage{multirow}
\usepackage{url}
\def\BibTeX{{\rm B\kern-.05em{\sc i\kern-.025em b}\kern-.08em
    T\kern-.1667em\lower.7ex\hbox{E}\kern-.125emX}}
\begin{document}

\title{Cross-Dataset Semantic Segmentation Performance Analysis: Unifying NIST Point Cloud City Datasets for 3D Deep Learning\\
\thanks{Location Based Services, PSCR NIST, 325 Broadway, Boulder, 80305, Colorado, United States}
}

\author{\IEEEauthorblockN{Alexander Nikitas Dimopoulos}
\IEEEauthorblockA{
\textit{Location Based Services PSCR} \\
\textit{National Institute of Standards and Technology}\\
Boulder CO, United States of America \\
alexander.dimopoulos@nist.gov \\
ORCID 0009-0001-8742-3940}
\and
\IEEEauthorblockN{Joseph Grasso}
\IEEEauthorblockA{\textit{Location Based Services PSCR} \\
\textit{National Institute of Standards and Technology}\\
Boulder CO, United States of America \\
joseph.grasso@nist.gov \\
ORCID 0009-0000-6075-2958}
}

\maketitle

\begin{abstract}
This study analyzes semantic segmentation performance across heterogeneously labeled point-cloud datasets relevant to public safety applications, including pre-incident planning systems derived from lidar scans. Using NIST's Point Cloud City dataset (Enfield and Memphis collections), we investigate challenges in unifying differently labeled 3D data. Our methodology employs a graded schema with the KPConv architecture, evaluating performance through IoU metrics on safety-relevant features. Results indicate performance variability: geometrically large objects (e.g. stairs, windows) achieve higher segmentation performance, suggesting potential for navigational context, while smaller safety-critical features exhibit lower recognition rates. Performance is impacted by class imbalance and the limited geometric distinction of smaller objects in typical lidar scans, indicating limitations in detecting certain safety-relevant features using current point-cloud methods. Key identified challenges include insufficient labeled data, difficulties in unifying class labels across datasets, and the need for standardization. Potential directions include automated labeling and multi-dataset learning strategies. We conclude that reliable point-cloud semantic segmentation for public safety necessitates standardized annotation protocols and improved labeling techniques to address data heterogeneity and the detection of small, safety-critical elements. 
\end{abstract}


\section{Introduction}

The rapid advancement of 3D sensing technologies has revolutionized indoor mapping and localization, particularly in public safety applications where first responders require precise spatial data for effective emergency response. Point cloud data, which captures detailed spatial information of indoor environments, has emerged as a crucial tool for creating accurate building representations. However, the unstructured nature of point clouds, combined with the complexity of indoor environments, presents significant challenges in data processing and interpretation. Machine learning (ML) models, particularly deep neural networks, have shown promising results in automated point-cloud analysis. Artificial intelligence (AI) and ML techniques have shown potential in various safety-related domains, from the use of multilayer perceptrons to understand building occupancy [1] to the use of deep learning for object detection to prevent collisions in autonomous vehicles [2]. Furthermore, the ability to interpret detailed environmental sensor data is fundamental for the development of advanced safety tools, such as situation-aware evacuation assistance systems [3], highlighting the critical role of automated analysis. However, the effectiveness of these models is highly dependent on the quality and consistency of training data, especially in the demanding context of public safety.

Recent years have seen the development of several large-scale point-cloud datasets, with notable contributions such as SemanticKITTI [4] or Stanford’s 2D-3D-Semantics [5] establishing benchmarks for 3D semantic segmentation. However, data sets specifically focused on public safety applications remain limited, both in quantity and in uniformity. Furthermore, achieving reliable model performance across the diverse environments encountered by first responders requires training data that capture variations in building types, sensor data, and features. The NIST Point Cloud City (PCC) [6] dataset represents a focused effort to address this gap, providing detailed indoor scans specifically annotated for public safety applications. However, the utilization of such datasets faces a challenge when collections within them use different labeling methodologies, as inconsistent labeling can significantly impact model performance.

Building upon the challenge of inconsistent labeling previously mentioned, this research investigates a common scenario in the collection of public safety data. How effectively can semantic segmentation models learn when trained on datasets developed with the shared objective of identifying features relevant to emergency response, but created using different annotation methods? We explored this question through an analysis of the NIST Point Cloud City (PCC) dataset, specifically the Enfield and Memphis collections [7, 8]. Both collections aimed to annotate indoor environments for public safety applications, but were labeled by different organizations employing distinct protocols, leading to variations in the specific semantic classes chosen, their definitions, and annotation granularity. This required harmonization for a combined analysis. Understanding how to manage such heterogeneity is crucial for developing the comprehensive datasets needed to advance automated scene understanding tools, particularly pre-incident planning systems for first responders.

Our approach involves three key components: (1) developing a unified data format that harmonizes differently labeled point clouds, (2) implementing a pipeline that converts PCC data into a SemanticKITTI compatible structure for use with state-of-the-art models, and (3) conducting extensive experiments to evaluate model performance across datasets with varying labeling schema. This methodology enables us to quantitatively assess the impact of labeling inconsistencies on semantic segmentation performance, particularly for public safety features.

The primary contributions of this work include: (a) a systematic analysis of cross-dataset semantic segmentation performance in public safety-oriented point clouds, (b) an approach to unifying heterogeneously labeled datasets while preserving semantic meaning, and (c) quantitative insights into the impact of different labeling methodologies on model performance across various object classes relevant to public safety applications.

\section{Related Work}

\subsection*{Dataset Development and Standardization}

While point cloud collections like Stanford's 2D-3D-Semantics (indoor environments) and SemanticKITTI (autonomous driving) have advanced semantic segmentation capabilities for their respective domains, datasets specifically focused on public safety applications remain notably limited. These general purpose benchmarks, though valuable for developing processing techniques, typically lack the specialized annotation schemes and public safety oriented features necessary for first responder applications. Existing standardization efforts by organizations such as OGC [9] and ISO [10] appropriately focus on broader geo-spatial data representations, acknowledging the nature of point clouds as unstructured geometric data that can be transformed into various 3D formats. The challenge lies not in the lack of point cloud standards, but rather in balancing the processing efficiency that makes point clouds valuable with the need for rich semantic annotations required for safety-critical applications. Although point clouds offer computational advantages for 3D data processing, they inherently lack flexible labeling and instance recording capabilities, features that would compromise their processing efficiency if directly integrated. This creates a fundamental tension between maintaining the performance benefits of point-cloud representations and implementing the detailed semantic annotation schemes necessary for emergency response applications.

\subsection*{Semantic Segmentation Methods}

Semantic point-cloud segmentation approaches have evolved from basic clustering methods to sophisticated deep learning architectures. Early unsupervised techniques relied on geometric features and clustering algorithms to segment point clouds into meaningful regions. These methods, while computationally efficient, often struggled with complex indoor environments and irregular object boundaries. Recent supervised learning approaches, including KPConv [11], have demonstrated superior performance by using deep neural networks to learn hierarchical features directly from point clouds. These methods can better handle the unstructured nature of point-cloud data while maintaining permutation invariance, a critical property for point-based learning.

\subsection*{Public Safety Applications}

The application of point-cloud processing in public safety contexts presents unique challenges and requirements not typically addressed in general-purpose solutions. Although traditional point cloud segmentation has focused on common indoor objects and architectural elements, public safety applications require reliable identification of public safety-oriented features and potential hazards. Although deep learning has shown promise in the use of sensor data for safety, for example, using deep convolutional neural networks to improve public safety through the detection of cracks in concrete bridges [12], the lack of specialized data sets focused on the needs of first responders within indoor environments has limited the development of robust solutions specifically tailored to their operational requirements. Our work with the Point Cloud City dataset represents one of the first large-scale efforts to address this gap, providing annotated data specifically focused on safety-critical features relevant to emergency response.

\subsection*{Position in Current Literature}

Although the field has established robust point-cloud benchmarks for specific domains such as autonomous driving (e.g. SemanticKITTI) and indoor scene understanding (e.g. ScanNet [13], S3DIS [14]), our work addresses a significant gap in datasets specifically developed for emergency response applications. Existing collections effectively support semantic segmentation within their respective domains, but do not adequately represent the unique public safety oriented features required by first responders. This research bridges disparate labeling approaches across heterogeneous datasets focused on public safety, acknowledging the inherent tension between maintaining point-cloud processing efficiency and incorporating rich semantic annotations. By developing a graded schema compatible with state-of-the-art architectures, we establish a framework for unified representation of public safety features that balances technical performance with practical emergency response requirements.

\section{Methodology}

\subsection{Data Pre-processing and Format Unification}

The Point Cloud City (PCC) dataset comprises annotated 3D indoor point clouds from multiple locations. This study uses collections from Enfield, Connecticut, and Memphis, Tennessee. An additional collection within PCC, the Hancock County Point Cloud City dataset from Mississippi [15], was excluded. This exclusion was necessary because its fundamentally different annotation structure, developed independently for that project, proved incompatible with the unification methodology applied to the Enfield and Memphis data used here. The Enfield collection includes various buildings, such as schools, administrative facilities, and university structures, developed through collaboration between the Enfield Fire Department and researchers at the University of Connecticut. The Memphis collection (``Map 90'') covers substantial indoor spaces in multiple facilities, captured using lidar-equipped backpacks, 360 ° cameras and environmental sensors, with time-synchronized images and environmental data.

To facilitate compatibility with state-of-the-art 3D segmentation models, we adopted the SemanticKITTI dataset structure, which provided advantages including simple point-cloud representation, compact binary files, well-documented folder structure, and clear label generation methodology. This format enabled direct integration with the Open3D-ML [16] software pipeline and application of the KPConv model architecture.

Our model deployment used Open3D-ML with PyTorch [17] to train and evaluate KPConv on pre-processed datasets. KPConv was selected for its deformable convolution mechanisms that process point clouds directly without voxelization, making it well-suited for the irregular point densities in our heterogeneously collected datasets.

To enable cross-dataset analysis, the different labeling schemes used in the Enfield and Memphis datasets required unification. We developed a mapping, termed PCC-SKITTI for this project, which assigns consistent numerical identifiers to semantic classes derived from the original annotations. This process involved reconciling different naming conventions and addressing inconsistencies where original classes required interpretation or aggregation. Detailed mapping tables that illustrate this process are available in the project’s online repository [18]. This unification mapping prioritized public safety oriented features while maintaining compatibility with the selected model architecture.

\subsection{Pipeline Architecture}

The pipeline was implemented using Open3D-ML which supports the KPConv deep learning architecture. The technical conversion pipeline includes initial data loading with specialized routines for handling varying input formats, format standardization to ensure SemanticKITTI compatibility, and quality control measures to maintain data integrity throughout the process.

This comprehensive methodology established a standardized framework that maintains the integrity of the original data sets while allowing seamless integration with contemporary deep learning architectures. The resulting unified format facilitates both the evaluation of existing models and the development of new approaches to understanding the indoor scene and public safety applications.

\section{Experimental Setup}

This section outlines our evaluation framework for semantic segmentation performance on indoor point cloud data for public safety applications. We present our training configurations across multiple datasets, justify our architecture selection based on technical requirements for emergency response scenarios, and detail our implementation specifics including hardware, software frameworks, and dataset partitioning strategies. Our methodology focuses on assessing both model performance and cross-dataset generalization capabilities to identify critical safety characteristics essential to first responders.

\subsection{Training Configurations}

To assess cross-dataset compatibility and model adaptability, we implemented three distinct training configurations:

\begin{enumerate}
    \item \textbf{Combined Dataset}: Training on the complete unified PCC-SKITTI dataset (merged Enfield and Memphis collections)
    \item \textbf{Enfield-Only}: Training exclusively on the Enfield subset 
    \item \textbf{Memphis-Only}: Training exclusively in the Memphis subset
\end{enumerate}

This approach allowed us to quantify both individual dataset performance and cross-dataset generalization capabilities. The specific data splits used for each configuration are defined within the project's online code repository [18].

\subsection{Architecture Selection}

We selected KPConv as our primary architecture for three key reasons:

\begin{enumerate}
    \item \textbf{Deformable Convolution Mechanism}: Adaptability to varying point densities observed both between datasets and within individual scans
    \item \textbf{Direct Point Processing}: Ability to process raw point cloud data without voxelization, preserving spatial fidelity critical for public safety oriented feature identification
    \item \textbf{Permutation Invariance}: Consistent results regardless of point order, ensuring reliable performance across different scanning patterns and collection methodologies
\end{enumerate}

\subsection{Implementation Details}

All experiments were carried out using the PyTorch implementation of KPConv through the Open3D-ML framework (v0.13) on Ubuntu 20.04 with an NVIDIA A100 GPU. The selection of hyperparameters was guided by the settings validated in the original KPConv study. Specifically, we implemented values closely aligned with their recommendations for optimal performance on large-scale point-cloud tasks, including an initial learning rate of 0.01, momentum of 0.98, weight decay of 0.001, and a learning rate decay factor of 0.98477. The batch size was set to 1 primarily due to GPU memory constraints when processing the large point cloud scenes. The complete configuration is specified in the YAML file available in the PointCloudCity-Open3D-ML Github repository [18].

The data set splits followed a structured approach with sequence-based division:
\begin{itemize}
    \item Combined dataset: 87 training sequences (00-58, 118-133), 1 validation sequence (59), and 76 test sequences (60-117, 134-149)
    \item Enfield-only: 59 training sequences (00-58), 1 validation sequence (59), and 60 test sequences (60-118) 
    \item Memphis-only: 15 training sequences (119-133), 1 validation sequence (134), and 15 test sequences (135-149)
\end{itemize}

These sequence assignments distribute scans of the various types of building present in the Enfield and Memphis collections through the training, validation, and test partitions for each configuration.

\section{RESULTS AND ANALYSIS}

This section presents the performance evaluation of the KPConv semantic segmentation model across the unified Point Cloud City (PCC) dataset, comprising the distinct Enfield and Memphis collections. Using standard evaluation metrics, primarily Intersection over Union (IoU), this analysis assesses model effectiveness under different training configurations (Combined PCC, Enfield-only, Memphis-only).

\begin{table}[htbp]
\caption{Per-Class IoU and Accuracy Comparison Across Datasets. Best results per class and metric are highlighted in bold.}
\label{tab:filtered_safety_feature_iou_accuracy_v1_modified} 
\centering
\scriptsize 
\setlength{\tabcolsep}{3pt} 
\renewcommand{\arraystretch}{1.1} 
\begin{tabular}{|c|c|c|c|c|}
\hline
\textbf{Class} & \textbf{Metric} & \textbf{Enfield} & \textbf{Memphis Model} & \textbf{PCC} \\
\hline
\multirow{2}{*}{Unassigned}
 & IoU      & \textbf{0.474} & 0.013 & 0.094 \\ \cline{2-5}
 & Accuracy & \textbf{0.575} & 0.026 & 0.148 \\ \hline
\multirow{2}{*}{Door}
 & IoU      & 0.394 & 0.120 & \textbf{0.475} \\ \cline{2-5}
 & Accuracy & \textbf{0.804} & 0.129 & 0.622 \\ \hline
\multirow{2}{*}{Fire door} 
 & IoU      & 0.103 & 0.217 & \textbf{0.220} \\ \cline{2-5}
 & Accuracy & 0.172 & \textbf{0.869} & 0.478 \\ \hline
\multirow{2}{*}{Window}
 & IoU      & \textbf{0.217} & 0.000 & 0.189 \\ \cline{2-5}
 & Accuracy & 0.263 & 0.000 & \textbf{0.278} \\ \hline
\multirow{2}{*}{Exit sign} 
 & IoU      & \textbf{0.005} & 0.000 & 0.000 \\ \cline{2-5}
 & Accuracy & \textbf{0.011} & 0.000 & 0.000 \\ \hline
\multirow{2}{*}{Emergency lighting} 
 & IoU      & \textbf{0.096} & 0.000 & 0.041 \\ \cline{2-5}
 & Accuracy & \textbf{0.203} & 0.000 & 0.047 \\ \hline
\multirow{2}{*}{AED}
 & IoU      & 0.000 & 0.000 & \textbf{0.030} \\ \cline{2-5}
 & Accuracy & 0.000 & 0.000 & \textbf{0.031} \\ \hline
\multirow{2}{*}{Hydrant}
 & IoU      & 0.186 & 0.000 & \textbf{0.289} \\ \cline{2-5}
 & Accuracy & \textbf{0.592} & 0.000 & 0.566 \\ \hline
\multirow{2}{*}{Gas shutoff} 
 & IoU      & \textbf{0.082} & 0.021 & 0.070 \\ \cline{2-5}
 & Accuracy & \textbf{0.146} & 0.060 & 0.114 \\ \hline
\end{tabular}
\end{table}

\begin{figure}[htbp] 
  \centering 
  \includegraphics[width=0.9\linewidth]{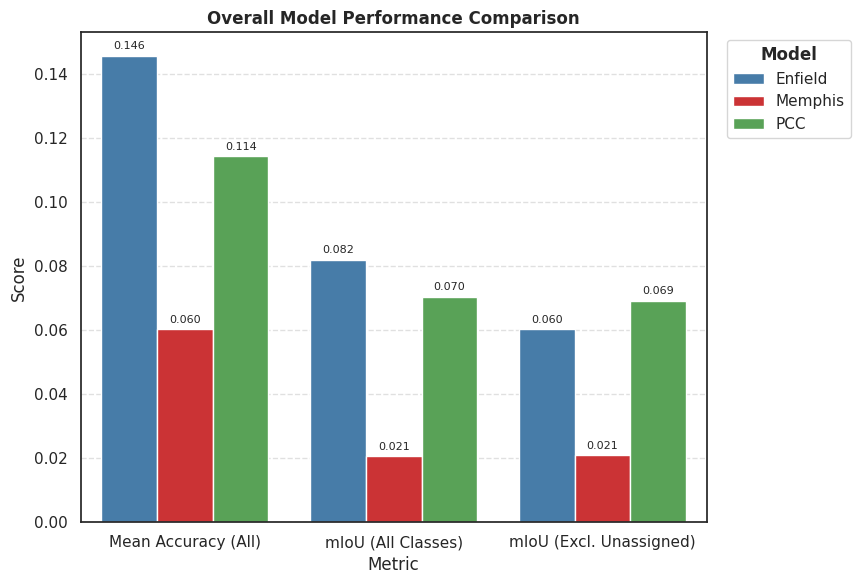}
  \caption{Model performance comparison for Enfield, Memphis, and combined PCC training. Metrics: Mean Accuracy (Mean Acc), mIoU (All classes), and mIoU (Excluding Unassigned).}
  \label{fig:overall_metrics} 
\end{figure}

\begin{table}[!ht]
\centering
\caption{Label Distribution Comparison (All Splits)}
\label{tab:label_distribution} 
\begin{tabular}{llrr}
\hline
\textbf{Label} & \textbf{Class Name} & \textbf{Enfield (\%)} & \textbf{Memphis (\%)} \\ \hline
0  & unassigned                   & 67.92 & 89.17 \\
1  & stairway                     &  1.46 &  0.12 \\
2  & door                         &  2.54 &  0.09 \\
3  & non exit door                &  9.60 &  6.87 \\
4  & fire door                    &  0.46 &  0.09 \\
5  & window                       &  7.45 &  2.04 \\
6  & roof access                  &  8.30 &  0.00 \\
7  & exit sign                    &  0.26 &  0.09 \\
8  & emergency lighting           &  0.08 &  0.01 \\
9  & smoke detector               &  0.11 &  0.55 \\
10 & extinguisher                 &  0.03 &  0.05 \\
11 & fire alarm                   &  0.71 &  0.17 \\
12 & person                       &  0.05 &  0.03 \\
13 & AED                          &  0.05 &  0.02 \\
14 & sprinkler                    &  0.08 &  0.01 \\
15 & standpipe                    &  0.01 &  0.00 \\
16 & utility shutoffs - electric  &  0.04 &  0.61 \\
17 & elevator                     &  0.15 &  0.08 \\
18 & hydrant                      &  0.00 &  0.00 \\
19 & gas shutoff                  &  0.70 &  0.00 \\ \hline
\vspace{1pt}
\end{tabular}
\small 
Comparative point label distributions for the Enfield and Memphis datasets, highlighting significant class imbalance inherent in this public safety data. 
\end{table}

\begin{figure*}[htbp] 
  \centering 
  \includegraphics[width=0.7\linewidth]{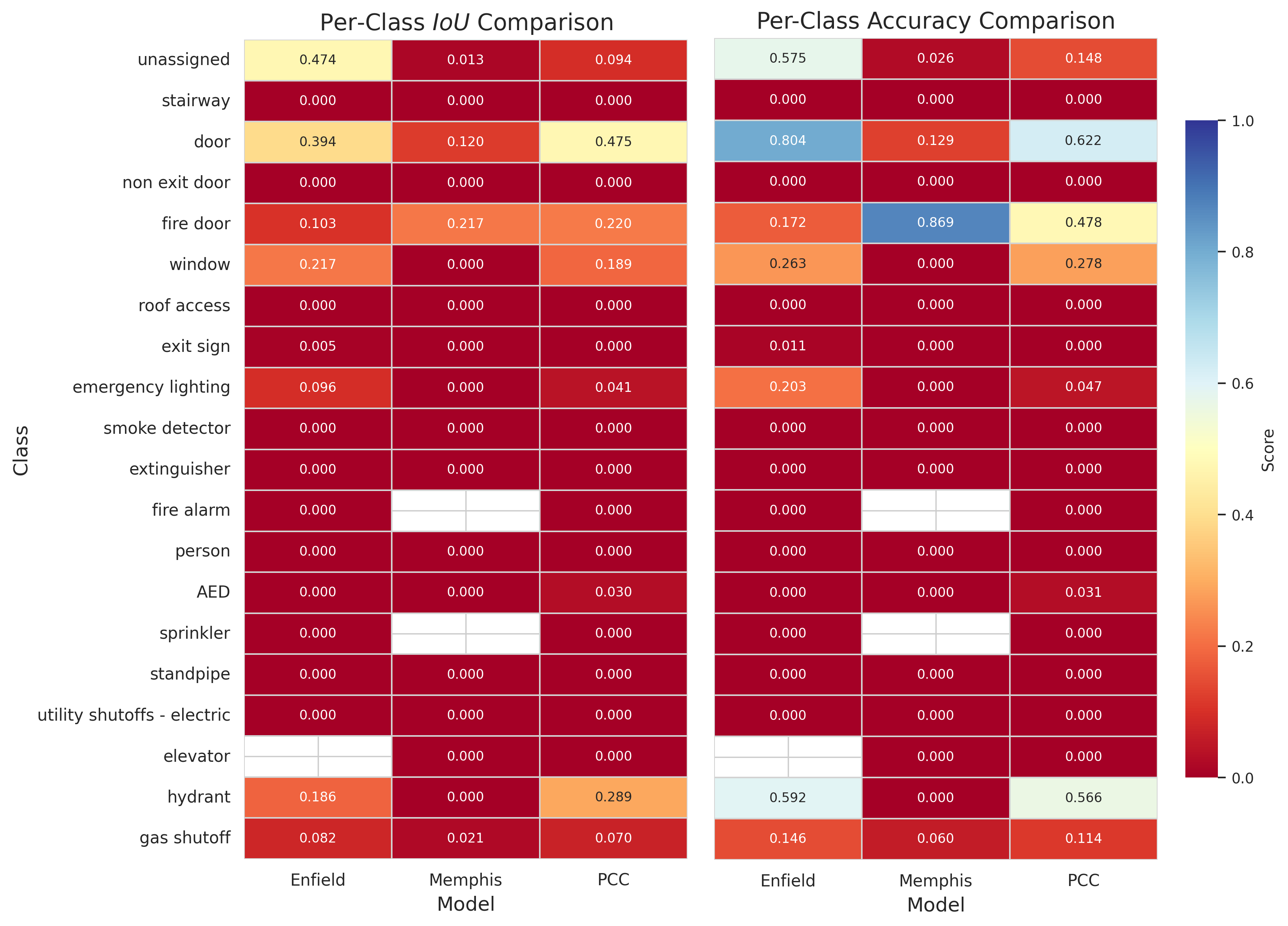}
  \caption{Heatmap comparison of per-class Intersection over Union (IoU) (left) and Accuracy (right) for the semantic segmentation model trained on different dataset configurations (Enfield, Memphis Model, PCC). Darker red indicates lower scores, while yellow/blue indicate higher scores.}
  \label{fig:iou_accuracy_heatmap}
\end{figure*}

\begin{figure}[htbp] 
  \centering 
\includegraphics[width=\linewidth]{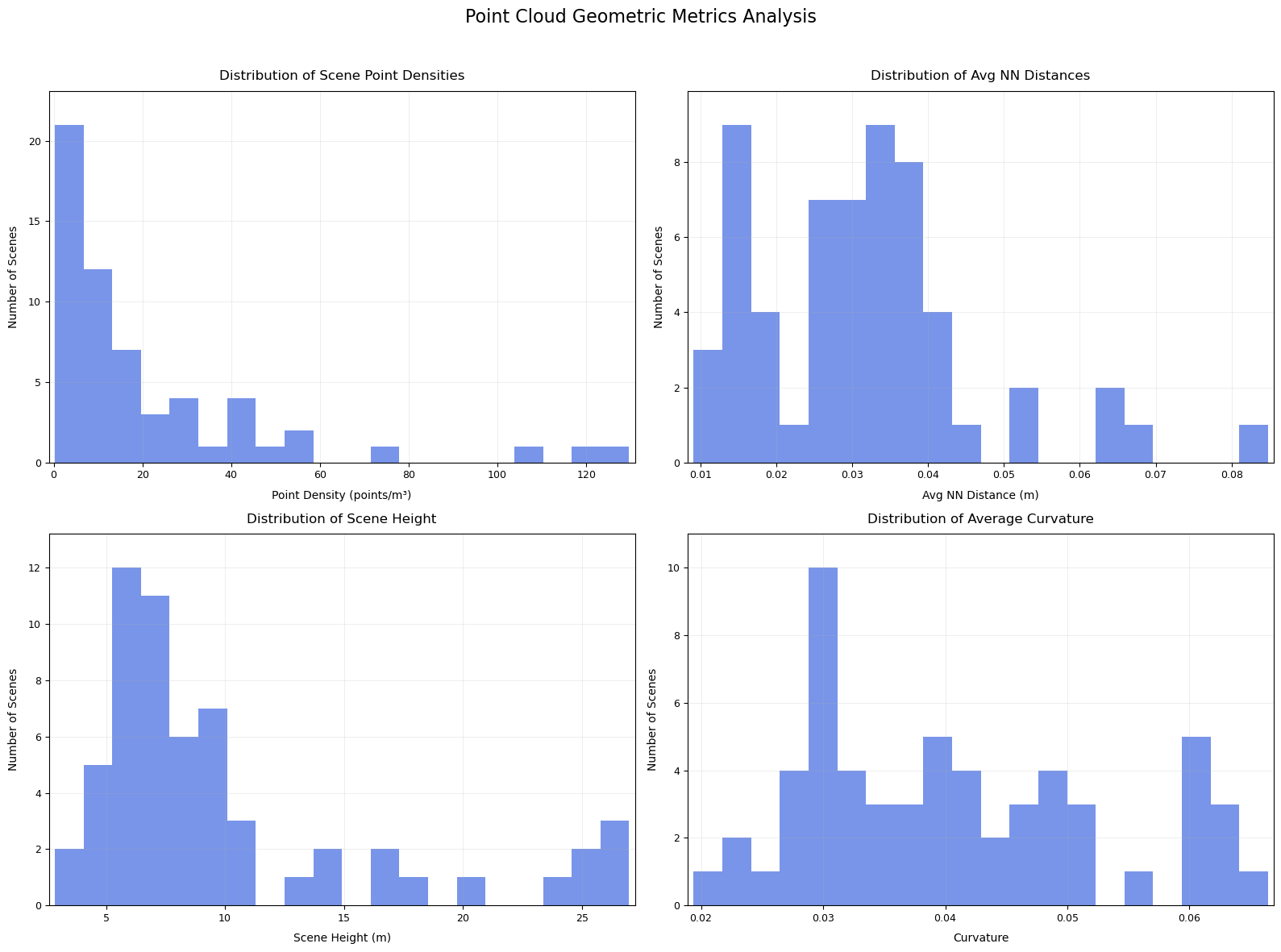}
  \caption{Fig. 3. Analysis of geometric metrics for the point cloud dataset. Histograms show distributions for: (Top Left) Point Density (points/m²), (Top Right) Average Nearest Neighbor (NN) Distance (m), (Bottom Left) Scene Height (m), and (Bottom Right) Average Curvature, characterizing the dataset's geometric properties.}
  \label{fig:pcc_pc_metrics} 
\end{figure}

As shown in Table I, the KPConv model performed effectively in segmenting several key classes, particularly those with distinct geometric characteristics or high prevalence in the data set. High performance was observed for architectural elements such as the 'door', which achieved high accuracy (0.804) with the Enfield training data and the best IoU (0.475) using the combined PCC dataset. Similarly, 'fire door' segmentation attained a high accuracy of 0.869 under the Memphis training configuration, achieving an IoU of 0.220 with the PCC model. Segmentation of the windows and the hydrants classes also yielded positive results, attaining the best IoU scores of 0.217 (Enfield) and 0.289 (PCC), respectively, indicating the model's capability to take advantage of the clearer geometric definitions present in these features. Furthermore, the model demonstrated consistently strong results for the 'unassigned' class, achieving the highest IoU (0.474) and accuracy (0.575) with the Enfield dataset, consistent with this class's dominance in point cloud data. These findings underscore the model's effectiveness in identifying larger, structurally defined elements within complex 3D point-cloud environments.

Semantic segmentation performance exhibited significant variation across the three training configurations evaluated: Enfield only, Memphis only, and the combined PCC dataset (Figure~\ref{fig:overall_metrics}). The model trained exclusively on the Enfield collection achieved superior overall results, yielding the highest mean accuracy (0.146) and mean intersection over the Union including all classes (mIoU All: 0.082). In stark contrast, the Memphis-only model demonstrated considerably lower performance across all metrics, attaining an mIoU (All) of only 0.021. Training on the combined PCC dataset generally resulted in performance metrics intermediate to the individual Enfield and Memphis models.

These divergent results underscore the profound impact of fundamental differences between the constituent datasets. Key contributing factors identified include disparate label distributions (Table II) and severe class imbalance inherent in realistic indoor scans, where public safety-critical features constitute a minute fraction ($<$1\%) compared to the predominant 'unassigned' points (70\%--90\%). Furthermore, varying geometric characteristics influenced the KPConv model's effectiveness; while proficient at segmenting large, structurally distinct elements like doors, it struggled significantly with compact public safety features lacking strong, unique geometric signatures. Although the combination of data sets aimed to improve generalization, it also introduced negative transfer effects, where the inclusion of highly imbalanced Memphis data degraded performance for specific classes compared to the Enfield-only model (Figure~\ref{fig:iou_accuracy_heatmap}).

This cross-dataset analysis illustrates the substantial challenges in training robust semantic segmentation models on heterogeneous point-cloud data derived from different sources and annotation protocols. The findings highlight limitations in reliably detecting small but vital safety features using current point-cloud segmentation approaches, particularly for public safety applications like emergency response pre-planning systems that demand high reliability across diverse real-world environments.

\section{Discussion and Conclusion}
Our cross-dataset analysis, utilizing the NIST Point Cloud City (PCC) datasets (Enfield and Memphis collections), reveals fundamental challenges in applying current 3D semantic segmentation models like KPConv to emergency response scenarios. The findings highlight a performance dichotomy: Although models effectively segment geometrically distinct architectural elements, they struggle significantly with the reliable identification of compact, public safety-oriented features vital for first responders.

An important observation is the correlation between segmentation performance and the geometric characteristics of object classes. Large structures with well-defined shapes (e.g., stairways, walls, doors, windows) are segmented with considerably higher reliability, as models like KPConv effectively leverage these robust geometric cues. However, this reliance on geometry proves detrimental for compact public safety-oriented items. These elements often lack unique 3D geometric signatures at typical lidar capture scales, exhibit variable appearances, and represent a minuscule volume within the overall point cloud, leading to poor and inconsistent segmentation. This suggests a need for methods that can incorporate greater semantic detail, beyond pure geometry, to reliably identify these crucial public safety relevant objects.

This challenge is compounded by the class imbalance inherent in realistic indoor scans. In our data sets, unassigned points constitute the vast majority (70\%–90\%), while public safety elements represent a tiny fraction (often $<$1\%). This severe underrepresentation limits the model's ability to learn these crucial minority classes. Our results demonstrate that simply increasing the data volume does not guarantee improved performance for these classes and can even induce negative transfer effects. This underscores that for emergency response applications, data quality, label consistency, and representation balance are paramount. Our development of a unified data format and conversion pipeline offers a practical step towards improving cross-dataset compatibility.

While our approach focused on harmonizing labels through a unified schema, recent methods such as Point Prompt Training (PPT) [19] offer an alternative strategy. PPT utilizes prompt-based learning and language guidance to dynamically unify disparate label spaces and mitigate the negative transfer effects observed in naive dataset mixing. Exploring such prompt-based techniques presents a promising avenue for future work to better leverage diverse public safety datasets without direct label modification.

Although the reliable detection of small safety features remains a challenge that requires improved semantic understanding, the demonstrated proficiency of current models in segmenting larger, geometrically simpler objects holds potential. These models could be effectively used to broadly label architectural features such as walls, floors, doors, and stairs across large point clouds. This capability could lead to the creation of more semantically enriched maps where a greater proportion of points are assigned meaningful labels, reducing the dominance of the 'unassigned' category and providing valuable structural context, even if specific safety items require alternative or supplementary detection methods. Addressing the gap for public safety-oriented features while leveraging the strengths for architectural elements will be key to advancing 3D scene understanding for public safety applications.

\end{document}